\documentclass[sigconf]{acmart}
\usepackage{ bbold }
\usepackage{subcaption}
\usepackage{graphicx}
\usepackage[]{algorithm2e}
\usepackage{easyeqn}
\graphicspath{{./}}

\usepackage{amsmath,amsfonts,bm}

\def\rvx{{\mathbf{x}}}
\def\rvy{{\mathbf{y}}}

\def\vtheta{{\bm{\theta}}}

\def\vu{{\bm{u}}}

\def\vx{{\bm{x}}}
\def\vy{{\bm{y}}}
\def\vz{{\bm{z}}}

\def\evtheta{{\theta}}

\newcommand{\E}{\mathbb{E}}

\newcommand{\KLD}[2]{D_{\mathrm{KL}} \left(#1 \, || \, #2 \right)}

\AtBeginDocument{%
  \providecommand\BibTeX{{%
    \normalfont B\kern-0.5em{\scshape i\kern-0.25em b}\kern-0.8em\TeX}}}

\setcopyright{acmcopyright}

\copyrightyear{2020}
\acmYear{2020}
\setcopyright{iw3c2w3}
\acmConference[WWW '20]{Proceedings of The Web Conference 2020}{April 20--24, 2020}{Taipei, Taiwan}
\acmBooktitle{Proceedings of The Web Conference 2020 (WWW '20), April 20--24, 2020, Taipei, Taiwan}
\acmPrice{}
\acmDOI{10.1145/3366423.3380061}
\acmISBN{978-1-4503-7023-3/20/04}

\acmSubmissionID{sp1718}

\begin{document}

\title{NCVis: Noise Contrastive Approach for Scalable Visualization}

\author{Aleksandr	Artemenkov}
\email{a.artemenkov@skoltech.ru}
\affiliation{%
  \institution{Skolkovo Institute of Science and Technology}
  \streetaddress{Bolshoy Boulevard 30, bld. 1}
  \city{Moscow}
  \state{Russia}
  \postcode{121205}
}
\affiliation{%
  \institution{Moscow Institute of Physics and Technology}
  \streetaddress{Institutskiy pereulok 9}
  \city{Dolgoprudny}
  \state{Moscow Region}
  \country{Russia}
  \postcode{141701}
}

\author{Maxim	Panov}
\email{m.panov@skoltech.ru}
\affiliation{%
  \institution{Skolkovo Institute of Science and Technology}
  \streetaddress{Bolshoy Boulevard 30, bld. 1}
  \city{Moscow}
  \country{Russia}
  \postcode{121205}
}

\renewcommand{\shortauthors}{Artemenkov, Panov}

\begin{abstract}
  Modern methods for data visualization via dimensionality reduction, such as t-SNE, usually have performance issues that prohibit their application to large amounts of high-dimensional data. In this work, we propose NCVis -- a high-performance dimensionality reduction method built on a sound statistical basis of noise contrastive estimation. We show that NCVis outperforms state-of-the-art techniques in terms of speed while preserving the representation quality of other methods. In particular, the proposed approach successfully proceeds a large dataset of more than 1 million news headlines in several minutes and presents the underlying structure in a human-readable way. Moreover, it provides results consistent with classical methods like t-SNE on more straightforward datasets like images of hand-written digits. We believe that the broader usage of such software can significantly simplify the large-scale data analysis and lower the entry barrier to this area.
\end{abstract}




\maketitle

\section{Introduction}
  \begin{figure}[t]
    \includegraphics[width=0.46\textwidth]{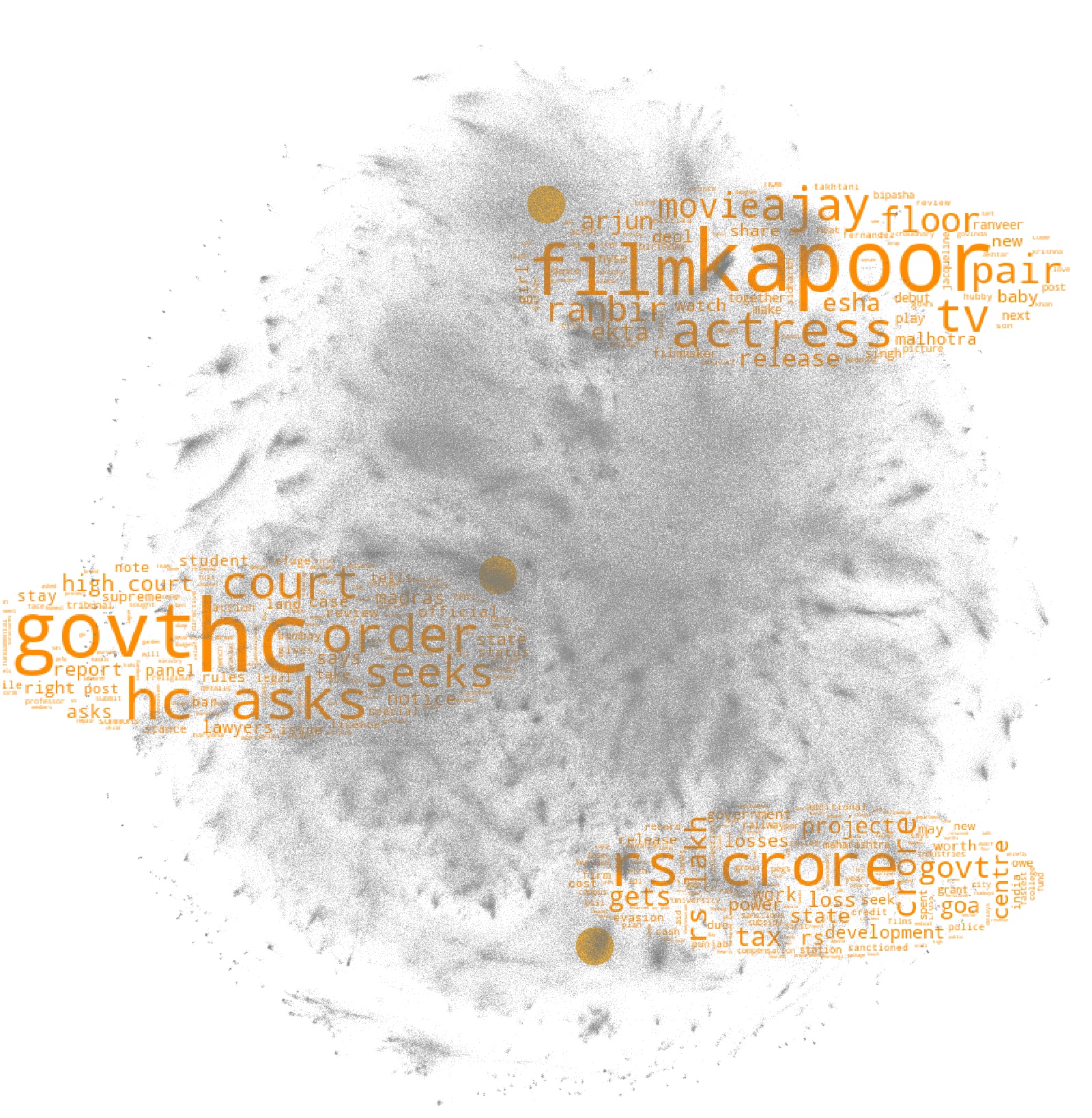}
    \caption{NCVis visualization of more than \textbf{2} million news headlines~\cite{NewsIndia} in less than \textbf{15} minutes. The word clouds illustrate the most frequent words used in the headlines from the area.}
  \end{figure}

  With the growth of the internet, extensive data sets became ubiquitous. However, it is not always feasible to tag every sample as there are too many of them. Even though we cannot perform a specific labeling procedure, this does not mean that there is no underlying structure in the data. On the contrary, such a structure is almost always present but requires a thorough study to be retrieved. The classical approach is to focus on a considerably smaller subset and try to find patterns there. But, intuitively, we can benefit from using the data, especially when one recalls the advances of statistical methods. 

  The classical linear approaches to dimensionality reduction, such as Principal Component Analysis (PCA; \cite{Hotelling1933,Jolliffe2011,Jolliffe2016}), are computationally efficient and widely used for data preprocessing and feature extraction. However, their linearity usually does not allow them to obtain high-quality low-dimensional representations that would be useful for visualization. Though the development of nonlinear dimensionality reduction methods such as Multidimensional scaling (MDS; \cite{Torgerson1952}), Isomap~\cite{Tenenbaum2000}, Locally Linear Embedding (LLE; \cite{Roweis2000}), Laplacian Eigenmaps~\cite{Belkin2002} and Local Tangent Space Alignment (LTSA; \cite{Zhang2004}) allowed to improve the quality of low-dimensional embeddings, their strong performance on artificial data is often not supported by comparable results on real-world high-dimensional data.

  The situation has changed when t-SNE~\cite{maaten2008visualizing} -- an approach focused on interpretable dimensionality reduction -- was introduced and was quickly considered to be a \textit{de-facto} standard for data visualization via dimensionality reduction. The computational performance of t-SNE is enough to provide a positive user experience with small to medium-sized data, but its poor scalability does not allow for processing large data sets. Several optimizations~\cite{van2014accelerating, linderman2019fast} were proposed to solve this problem, but none of them were able to provide a considerable improvement of its performance. The recently introduced LargeVis~\cite{tang2016visualizing} and Umap~\cite{mcinnes2018umap} focus on the usage of stochastic optimization for the objective function similar to the one of t-SNE. Both these methods achieve a considerable speedup as compared to their competitors, but still are based on a large set of heuristics, which results in difficulties with understanding the underlying idea and practical usage.

  In this paper, we formulate a statistical approach to design high-performance dimensionality reduction algorithms. In this case, the need for heuristics is minimized, so the efficacy can be granted for free. Our approach is based on the theory of Noise Contrastive Estimation (NCE; \cite{gutmann2010noise, gutmann2012noise}) and proceeds from the assumption that though many representations may have structural similarities with the target object, only a few of them avoid including unnecessary details. This variant of Occam's razor can be interpreted in the following way: while maximizing the correspondence with real data, also minimize the similarity with noise.

  When applied to the dimensionality reduction problem, NCE approach allows us to introduce Noise Contrastive Visualization (NCVis)~~--- a highly-scalable method for visualization via dimensionality reduction. The main issue which prevents the efficient batch training of t-SNE is the necessity to re-normalize the distribution in embedding space at every step. NCVis allows to avoid this issue and its scalability becomes the direct consequence of self-normalizing properties of NCE~\cite{gutmann2010noise, gutmann2012noise}. 

  Our main contributions can be summarized as follows:
  \begin{itemize}
    \item We propose a principled statistical approach \textit{NCVis} to obtain scalable dimensionality reduction algorithms basing on Noise Contrastive Estimation methodology. The approach allows for efficiently parallelizable batch training of embeddings.

    \item We implement NCVis as a multi-platform Python package\footnote{https://github.com/stat-ml/ncvis} providing a stable out-of-the-box experience.

    \item Experiments have proved that NCVis outperforms state-of-the-art methods for visualization via dimensionality reduction in terms of speed without losing a high visualization quality.

    \item Due to the parallel nature of the algorithm, NCVis gains significant benefits from hardware leveraging which allows us to visualize millions of high-dimensional objects in several minutes using a standard multi-core PC.
  \end{itemize}

\section{Problem Statement}
  In this paper, we address the problem of building a low-dimensional representation of high-dimensional data:
  \begin{EQA}[c]
  	\bigl\{\widehat{\vz}_i \in \mathbb{R}^{M}\bigr\}_{i=1}^N \to \bigl\{\vz_i \in \mathbb{R}^{m}\bigr\}_{i=1}^N, \text{ where }m \ll M.
  \end{EQA}
  We also suppose that some distance function $d(\cdot, \cdot)\colon \mathbb{R}^{M} \times \mathbb{R}^{M} \to [0, +\infty)$ can be defined for each pair of vectors from the set $\{\widehat{\vz}_i\}_{i=1}^N$. We assume that the value of this function should increase as the dissimilarity between the points goes up. That is why there is no demand for metric properties except for non-negativity: $d(\cdot, \cdot) \ge 0$.

\section{Noise Contrastive Visualization}
  In this paper, we introduce Noise Contrastive Visualization (NCVis)~--- a highly-scalable method for visualization via dimensionality reduction based on the Noise Contrastive Estimation (NCE) approach. Its implementation is based on a special optimization procedure, which makes it well suited for parallelization.

  An informal overview of the method goes as follows. One can select pairs of neighbors from the original high-dimensional data set. Suppose, that this is done in random manner: the higher the distance between points, the lower the probability for them to be selected. Now suppose that we also given a low-dimensional representation of the original data points, so we can perform a similar procedure there. It seems reasonable, that a good representation should induce the same pairs that were observed in the original data. Note that in order to build such a representation, we need to compute all the pairwise distances in the embedding. This will allow us to compare them and use terms ``large'' and ``small'' properly. However, we can reformulate the objective. Consider noisy representation where all the neighbors probabilities are almost equal. We can now use them as reference to mark lower-dimensional probabilities as either large or small. In this case we do not need to gather information about all the pairs of neighbors, it will be enough to know the scale of lower-dimensional probabilities with respect to the noisy probabilities. 

\subsection{Noise Contrastive Estimation}
  We start by describing NCE approach and establishing the notations to be used further. Noise Contrastive Estimation (NCE) was introduced in~\cite{gutmann2010noise} and developed in a more recent work~\cite{gutmann2012noise}. Let~$P_d$ be a data distribution with density~$p_d$ and samples~$X = \{\vx_i\}_{i=1}^{T_d}$ coming from this distribution. Consider also noise distribution~$P_n$ with density~$p_n$ and corresponding samples~$Y = \{\vy_i\}_{i=1}^{T_n}$. Unite the two sets: $ X \cup Y = U = \{\vu_i\}_{i=1}^{T_d+T_n}$, and assign a binary label $C_t$ to each element $\vu_t$: $C_t = 1$ if $\vu_t \in X$ and $C_t = 0$ if $\vu_t \in Y$. We also define noise ratio as $\nu = \frac{T_n}{T_d}$. Our initial goal is to obtain an approximation of~$p_d$ with some model~$p_m(\cdot; \vtheta)$. It is important to notice that normalization is included in the parameters $\vtheta$, so, generally speaking, the model is unnormalized. NCE objective $J_T(\vtheta)$ is then given by the principle of maximum likelihood:
  \begin{EQA}[c]
    J_T(\vtheta) \to \max_{\vtheta},
  \end{EQA}
  where
  \begin{EQA}
    J_T(\vtheta)
    &=&
    \frac{1}{T_d}\Big\{ \sum_{i=1}^{T_d} \log \frac{p_m(\vx_i; \vtheta)}{p_m(\vx_i;\vtheta) + \nu p_n(\vx_i)} \\
    &+&
    \sum_{i=1}^{T_n} \log \frac{\nu p_n(\vy_i)}{p_m(\vy_i;\vtheta) + \nu p_n(\vy_i)}\Big\}.
  \end{EQA}
  In order to apply the law of large numbers, we can also rewrite it as
  \begin{EQA}
    J_T(\vtheta)
    & = 
    \frac{1}{T_d}\sum_{i=1}^{T_d} \log \frac{p_m(\vx_i; \vtheta)}{p_m(\vx_i;\vtheta) + \nu p_n(\vx_i)} \\
    &+ \nu\frac{1}{T_n}\sum_{i=1}^{T_n} \log \frac{\nu p_n(\vy_i)}{p_m(\vy_i;\vtheta) + \nu p_n(\vy_i)}.
  \end{EQA}
  Assume that both $P_d$ and $P_n$ have a finite mean. When $T_d$ tends to infinity (and $T_n$ also tends to infinity as $T_n = \nu T_d$) the following objective is obtained using the weak law of large numbers:
  \begin{EQA}
  \label{eq:J_nce}
    J(\vtheta)
    & = &
    \E_{\rvx\sim P_d} \left[ \log \frac{p_m(\rvx; \vtheta)}{p_m(\rvx;\vtheta) + \nu p_n(\rvx)} \right] \\
    & + &
    \nu \E_{\rvy\sim P_n}\left[\log \frac{\nu p_n(\rvy)}{p_m(\rvy;\vtheta) + \nu p_n(\rvy)}\right].
  \end{EQA}
  Due to a stochastic nature of the objective function, the first order batch optimization methods can be beneficial in the case of the large number of parameters $\vtheta$. The explicitly written gradient with respect to the vector of parameters is given by
  \begin{EQA}[c]
  \label{eq:J_nce_grad}
   \frac{\partial}{\partial \evtheta_i} J(\vtheta) = \int \frac{p_d(\vx)  - p_m(\vx; \vtheta)}{1 + \frac{p_m(\vx; \vtheta)}{\nu p_n(\vx)}} \frac{\partial}{\partial \evtheta_i} \log p_m(\vx; \vtheta) d \vx .
  \end{EQA}
  
  A strong resemblance can be noticed between the proposed approach and a standard ML estimate for unnormalized models. Maximum Likelihood objective function $L(\vtheta)$ and its gradient are given by  
  \begin{EQA}[c]
    L(\vtheta) = \E_{\vx \sim P_d} \left[ \log \frac{p_m(\vx; \vtheta)}{\int p_m(\vy;\vtheta)d \vy} \right],
    \\
    \frac{\partial}{\partial \evtheta_i} L(\vtheta) = \int \left\{ p_d(\vx) - \frac{p_m(\vx; \vtheta)}{\int p_m(\vy;\vtheta)d \vy}\right\} \frac{\partial}{\partial \evtheta_i} \log p_m(\vx;\vtheta) d \vx.
    \label{eq:L_grad}
  \end{EQA}

  The similarity between Equation~\eqref{eq:J_nce_grad} and Equation~\eqref{eq:L_grad} goes even further, when noise ratio $\nu$ tends to infinity:
  \begin{EQA}[c]
    \frac{\partial}{\partial \vtheta}J(\vtheta) \underset{\nu \to \infty}{\longrightarrow} \int \left( p_d(\vx) -p_m(\vx; \vtheta) \right) \frac{\partial}{\partial \vtheta} \log p_m(\vy;\vtheta),
  \end{EQA}
  which coincides with Equation~\eqref{eq:L_grad} if $\int p_m(\vx;\vtheta) d \vx = 1$. As it follows, every stationary point of MLE objective where $p_m(\vx;\vtheta)$ is normalized is also a stationary point of NCE objective and \textit{vice versa}. 

  Recall that t-SNE~\cite{maaten2008visualizing} minimizes Kullback-Leibler divergence between $p_d$ and $p_m$, and thus maximizes the corresponding likelihood. The optimum is usually searched for by a gradient descent or other first-order optimization methods. To conclude, if noise ratio $\nu$ is high enough and the resulting model is normalized, there is no significant difference between using t-SNE and NCVis (see also the experimental validation in Section~\ref{sec:quality}).

\subsection{Probabilistic Model}
\label{sec:prob}
  In this section, we specify the particular instance of the approach described above to provide a simple but powerful probabilistic method for obtaining high-quality visualizations.

  First of all, we reintroduce the main idea of t-SNE which starts from defining conditional probabilities $\mathbb{P} =\{p_{ij}\}_{i, j = 1}^N$ of high-dimensional points:
  \begin{EQA}[c]
  \label{eq:tsne_p}
    p_{ij} = \frac{\widehat{p}_{ij}}{\sum_{k \neq l}\widehat{p}_{kl}}, \quad
    \widehat{p}_{ij} = 
    \exp \left[-\frac{\|\widehat{\vz}_i - \widehat{\vz}_j\|_2^2}{2\sigma_i^2}\right].
  \end{EQA}
  In Equation~\eqref{eq:tsne_p} the variances $\{\sigma_i^2\}_{i=1}^N$ are selected to keep the entropy of distributions $\{p_{ij}\}_{j=1}^N$ close to the predefined value for each $i$. Similarly, for points in low-dimensional space the probabilities $\mathbb{Q} =\{q_{ij}\}_{i, j = 1}^N$ are introduced:
  \begin{EQA}[c]
    q_{ij} = \frac{\widehat{q}_{ij}}{\sum_{k \neq l} \widehat{q}_{kl}}, \quad
    \widehat{q}_{ij} = 
    \frac{1}{1 + \|\vz_i - \vz_j\|_2^2}.
  \end{EQA}  

  Finally, the low-dimensional embeddings $\{\vz_i\}_{i=1}^N$ are obtained by minimizing Kullback-Leibler divergence between the distributions:
  \begin{EQA}[c]
    \KLD{\mathbb{P}}{\mathbb{Q}} \to \min_{}.
  \end{EQA}
  The main problem here is that every $q_{ij}$ is normalized, which requires the computation of aggregate sum $\sum_{k \neq l}\widehat{q}_{kl}$ and leads to a significant slowdown and parallelization issues. This difficulty can be overcome by switching to unnormalized models which can be done by using NCE since it provides a robust framework to handle such problems.

  We note, that probabilities $p_{ij}$ from~\eqref{eq:tsne_p} quickly become negligible when the distance between the points increases. Having taken it into account, we propose to simply use neighborhood indicators instead of Gaussian type probabilities in~\eqref{eq:tsne_p}. For a fixed number neighbours $k$ we define
  function 
  \begin{EQA}[c]
    \mathbb{1}_i^k(j) =
    \begin{cases}
      1, & \text{if point $\widehat{\vz}_j$ is among $k$ nearest neighbors of point $\widehat{\vz}_i$},\\
      0, & \text{otherwise or if $i = j$.}
    \end{cases}
  \end{EQA}
  Another function that is useful for us is $\mathbb{1}^k(i, j) = \mathbb{1}_i^k(j) \vee \mathbb{1}_j^k(i)$, which is non-zero if at least one of the points is among $k$~nearest neighbors of the other point. Now for each pair of points $(i, j)$ in the high dimensional space we introduce the probability of observing their co-occurrence
  \begin{EQA}[c]
    p_{ij} = \frac{\mathbb{1}^k(i, j)}{\sum_{i \neq j} \mathbb{1}^k(i, j)}.
  \end{EQA}

  For a lower-dimensional space, unlike t-SNE, we propose to treat normalization constant $Q$ as one of the parameters:
  \begin{EQA}[c]
    q_{ij} = \widehat{q}_{ij} e^{-Q}, \quad \widehat{q}_{ij} = \left(1 + a\|\vz_i - \vz_j\|^{2b}\right)^{-1},
  \end{EQA}
  where $a$ and $b$ are some fixed values. Such an expression for probabilities is inspired by Umap~\cite{mcinnes2018umap}.

  Finally, using NCE notation, we observe the nearest neighbor graph
  \begin{EQA}[c]
    X \subseteq \{(i, j)\}_{i,j = 1}^{N} ~\text{ with edge probabilities }~ p_{d}(i, j) = p_{ij}.
  \end{EQA}

  As proposed before, we are intended to unite the normalization parameter $Q$ and vector representations $\{\vz_i\}_{i=1}^N$ into a single vector parameter $\vtheta = \{Q, \{\vz_i\}_{i=1}^N\}$. In this notation the modelling distribution is given by 
  \begin{EQA}[c]
    p_{m}(i, j; \vtheta) = q_{ij}.
  \end{EQA}
   
  To be consistent in simplification, we propose the noise distribution with the following sampling procedure:
  \begin{enumerate}
    \item Sample edge $(i, j)$ according to distribution $\{p_d(i, j)\}_{i=1, j=1}^N$.
    \item Select $k$ from uniform distribution on $\{l\}_{l=1, l \neq i}^N$.
    \item Return edge $(i, k)$ as the sample. 
  \end{enumerate}
  This induces the following noise distribution:
  \begin{EQA}[c]
    Y = \{(i, j)\}_{i,j=1}^{N} ~\text{ with probabilities }~  p_{n}(i,j) = \frac{1}{N-1} \sum_{k=1}^N p_{ik}.
  \end{EQA}

  After the construction of the distributions, which one can easily sample from, we will now utilize stochastic gradient ascent to optimize NCE objective.

\subsection{Optimization Algorithm}
  The whole purpose of using Noise Contrastive Estimation is the ability to leverage batch optimization. The expectations in Equation~\eqref{eq:J_nce} can be approximated by sampling from the distributions $P_d$ and $P_n$, respectively. That is why it was essential to keep them as simple as possible. Otherwise, efficient parallelization may not be enough to compensate for the complexity of sampling. The details can be found in Algorithm~\ref{algo:embed}.

  \begin{algorithm}[t]
    \KwIn{A set $\{\widehat{\vz}_i \in \mathbb{R}^{M}\}_{i=1}^N$ of vectors}
    \KwOut{Visualization $\{\vz_i \in \mathbb{R}^{m}\}_{i=1}^N$ of the input}
    \BlankLine
    Construct $P_d$ and $P_n$ from $\{\widehat{\vz}_i\}_{i=1}^N$

    Initialize $\vtheta = \{Q, \{\vz_i\}_{i=1}^N\}$

    \For{$epoch$ $\gets 1$ \KwTo $N_{epochs}$ } {
      \For{$sample$ $\gets 1$ \KwTo $N_{samples}$} {
        $\vx \gets (v_1, v_2) \sim P_d$ \tcp*{sample real edges} 
        \For{$j$ $\gets 1$ \KwTo $\nu$} {
          $\vy_j \gets (v_1, v_2) \sim P_n$ \tcp*{sample noise} 
        }
        $J_T(\vtheta) \gets 0$;

        $J_T(\vtheta) \gets J_T(\vtheta) + \log \frac{p_m(\vx; \vtheta)}{p_m(\vx; \vtheta) + \nu p_n(\vx)}$;
        
        $J_T(\vtheta) \gets J_T(\vtheta) + \sum_{j=1}^{\nu} \log \frac{\nu p_n(\vy_j)}{p_m(\vy_j; \vtheta) + \nu p_n(\vy_j)}$;
        
        $\vtheta \gets \vtheta + \alpha \cdot \nabla J_T(\vtheta)$ \tcp*{gradient
        ascent step}
      }
    }
    \caption{Embedding construction algorithm.}
    \label{algo:embed}
  \end{algorithm}

  During the initialization phase, we use the Power Iteration method. The reason can be formulated as follows. If the nearest neighbor graph adjacency matrix has a block structure, the corresponding coordinates of the initialization vector will be located closely to each other. Thus, representations of the nodes from one community will be located near each other providing high values of likelihood. 

  In our setting, it is crucial to have an efficient solution for the nearest neighbor graph construction. To achieve it, we have selected the approach based on Hierarchical Navigable Small World graphs (HNSW), whose advantages over the alternatives are presented in detail in~\cite{malkov2018efficient}. The main idea of the method is to build a hierarchical structure of proximity graphs that allows localizing nearest neighbors via subsequent scale reduction efficiently.

\subsection{Complexity}
  If we use $k$ nearest neighbors for each point, then NCVis will consist of the following steps:
  \begin{enumerate}
    \item The nearest neighbor graph construction by HNSW has complexity $O(N \log N)$~\cite{malkov2018efficient}.

    \item Embedding initialization is performed with the Power Iteration method. Due to the sparsity of the graph, it has complexity $O(kN)$.

    \item Embedding optimization is a direct application of Algorithm~\ref{algo:embed} and thus also has complexity $O(kN)$.
  \end{enumerate}
  Consequently, the overall complexity is $O\bigl(N(k + \log N)\bigr)$ and is determined by the complexity of the graph construction approach for a sufficiently small number $k$ of nearest neighbors.

\section{Performance Evaluation}
  We compared the speed and visualization quality of NCVis to those of other existing methods. Our algorithm aims to provide state-of-the-art performance on large data sets while giving predictable output is standard cases. It is essential for a such software to show a good out-of-the-box performance and to have a convenient interface. The most popular methods in this category are Multicore t-SNE~\cite{Ulyanov2016}, Umap~\cite{mcinnes2018umap} and t-SNE implemented in Scikit-learn~\cite{scikitlearn}. However, we also want our method to be competitive with those which do not satisfy this criterion but are still efficient in practice such as FIt-SNE~\cite{linderman2019fast} and LargeVis~\cite{tang2016visualizing}. For all the experiments, we used NCVis with $15$ nearest neighbors, $50$ optimization epochs and the data set size as the number of samples.

\subsection{Speed Comparison}
\begin{figure}[t]
    \includegraphics[width=0.46\textwidth]{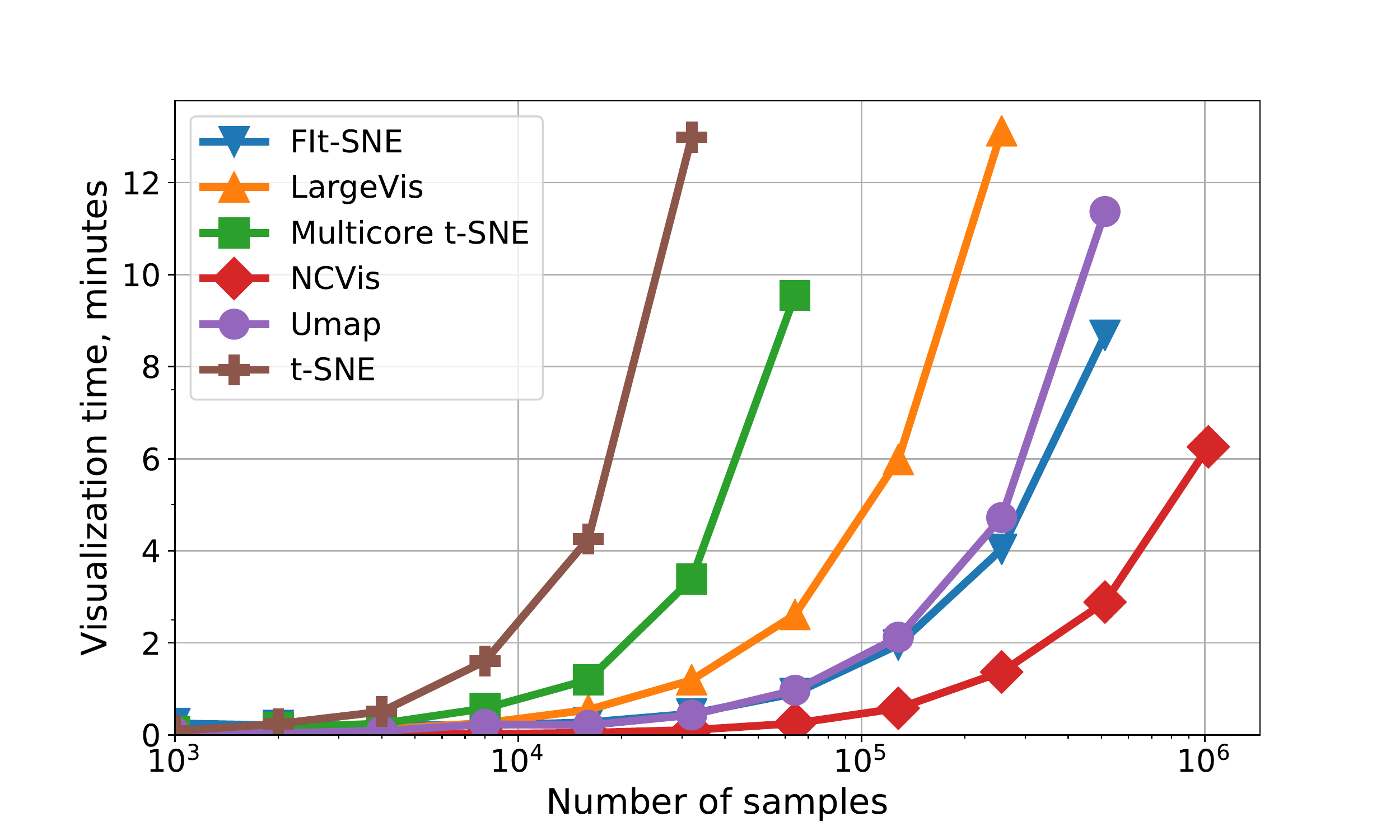}
    \caption{Comparison of the visualization methods by speed. Given same amount of time NCVis allows to process more than double number of samples compared to other methods, visualizing $10^6$ points in only $6$ minutes.}
    \label{fig:speed}
  \end{figure}

  We use preprocessed samples from the News Headlines Of India data set~\cite{NewsIndia} to perform the comparison. Test cases are generated by taking the first $1000, 2\cdot1000, \dots, 2^{10}\cdot1000$ samples from the data set. The hardware used is $12\,\times\,$$\text{Intel\textsuperscript{\textregistered}}$$\text{Core\textsuperscript{\texttrademark}}$  i7-8700K CPU @ 3.70GHz, $64\, \text{Gb RAM}$. Figure~\ref{fig:speed} summarized the results of the experiment which show that the performance of the proposed method is superior to its alternatives. It should be noted here that Scikit-learn t-SNE and Umap do not support parallelization while other methods benefit from it.

\subsection{Quality Comparison}
\label{sec:quality}
\begin{figure}[t]
    \centering
    \begin{subfigure}[b]{0.23\textwidth}
      \includegraphics[width=\textwidth]{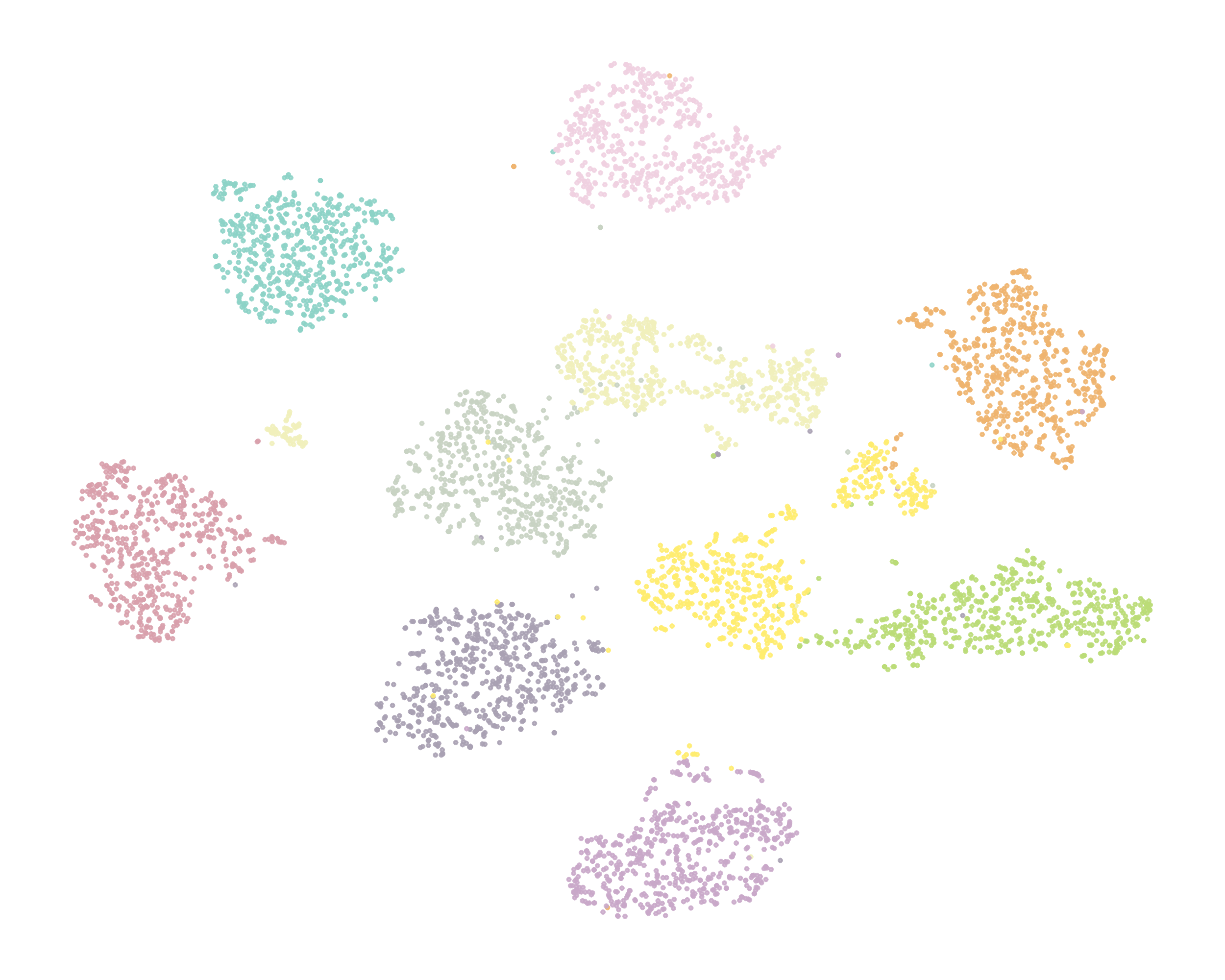}
      \subcaption{t-SNE, $29.5$s}
    \end{subfigure}
    \begin{subfigure}[b]{0.23\textwidth}
      \includegraphics[width=\textwidth]{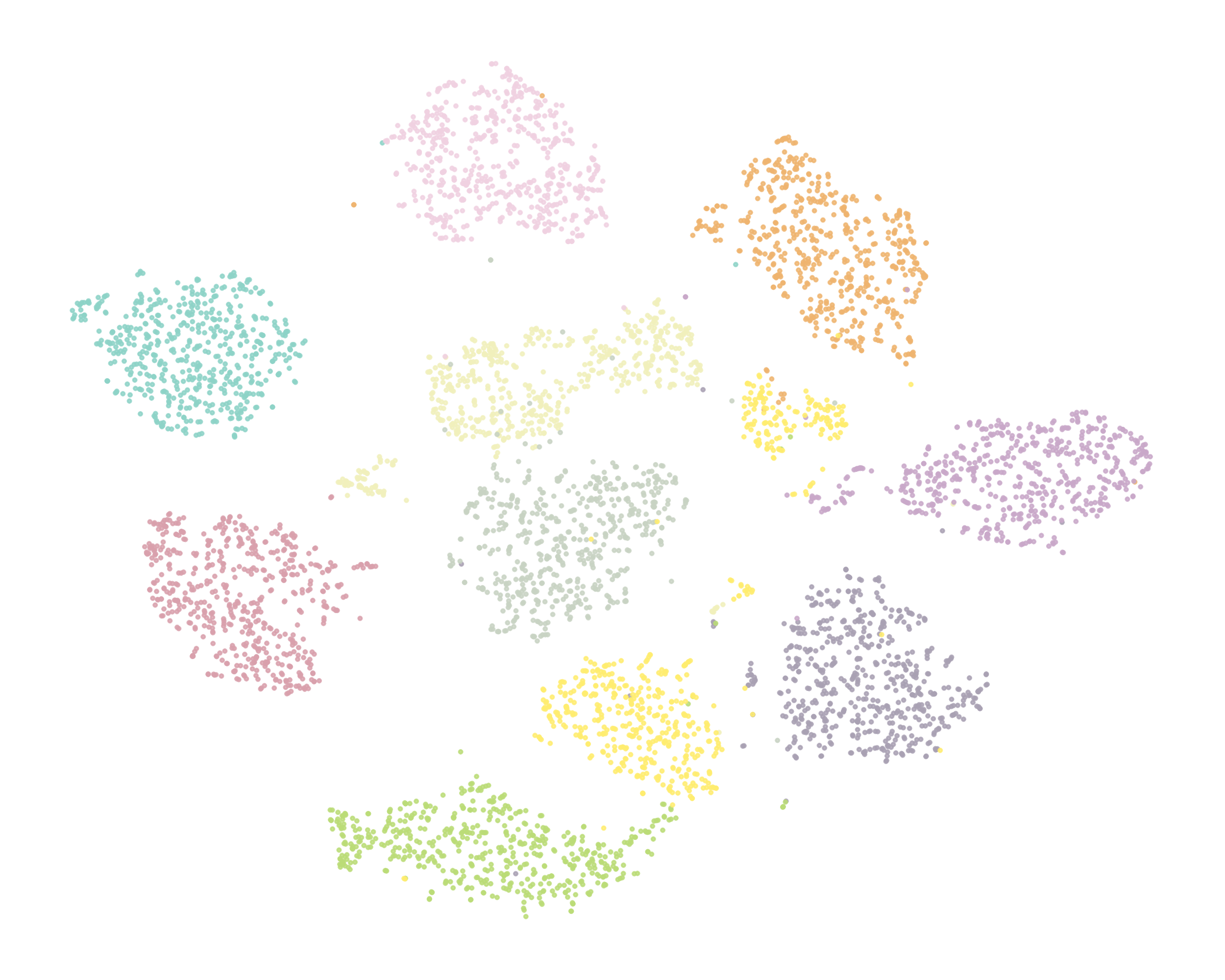}
      \subcaption{FIt-SNE, $17.4$s}
    \end{subfigure}\\
    \begin{subfigure}[b]{0.23\textwidth}
      \includegraphics[width=\textwidth]{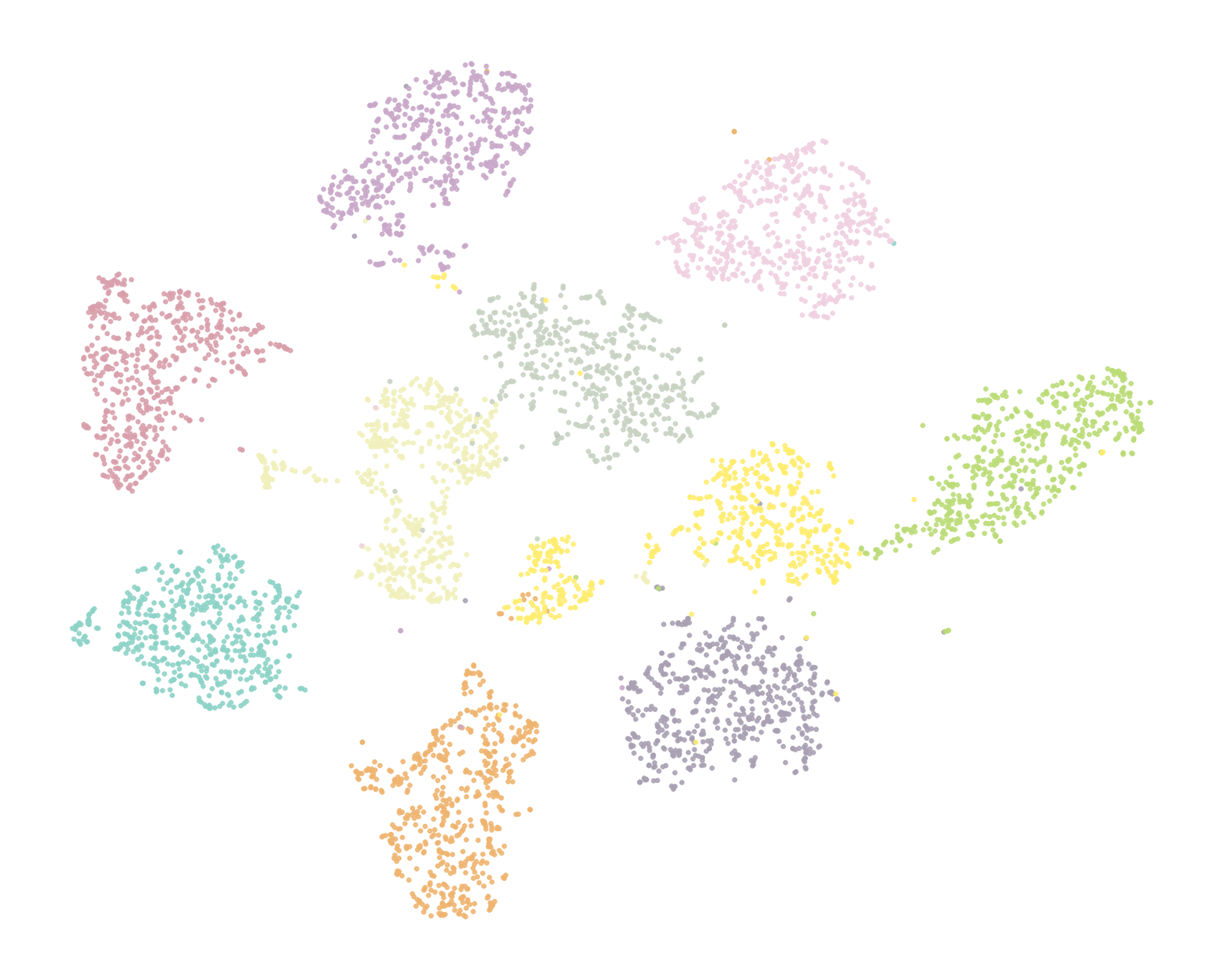}
      \subcaption{Multicore t-SNE, $14.3$s}
    \end{subfigure}
    \begin{subfigure}[b]{0.23\textwidth}
      \includegraphics[width=\textwidth]{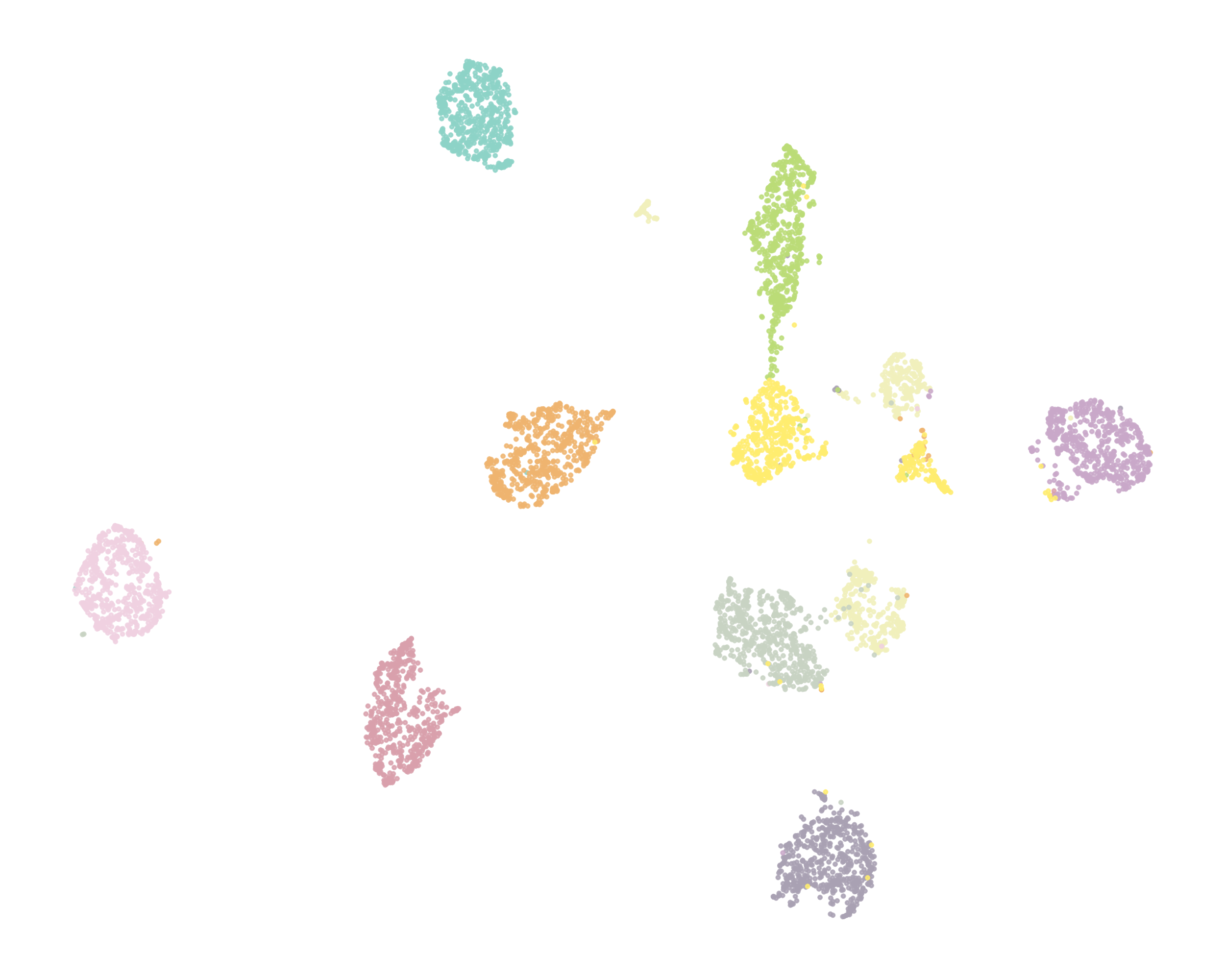}
      \subcaption{LargeVis, $9.7$s}
    \end{subfigure}\\
    \begin{subfigure}[b]{0.23\textwidth}
      \includegraphics[width=\textwidth]{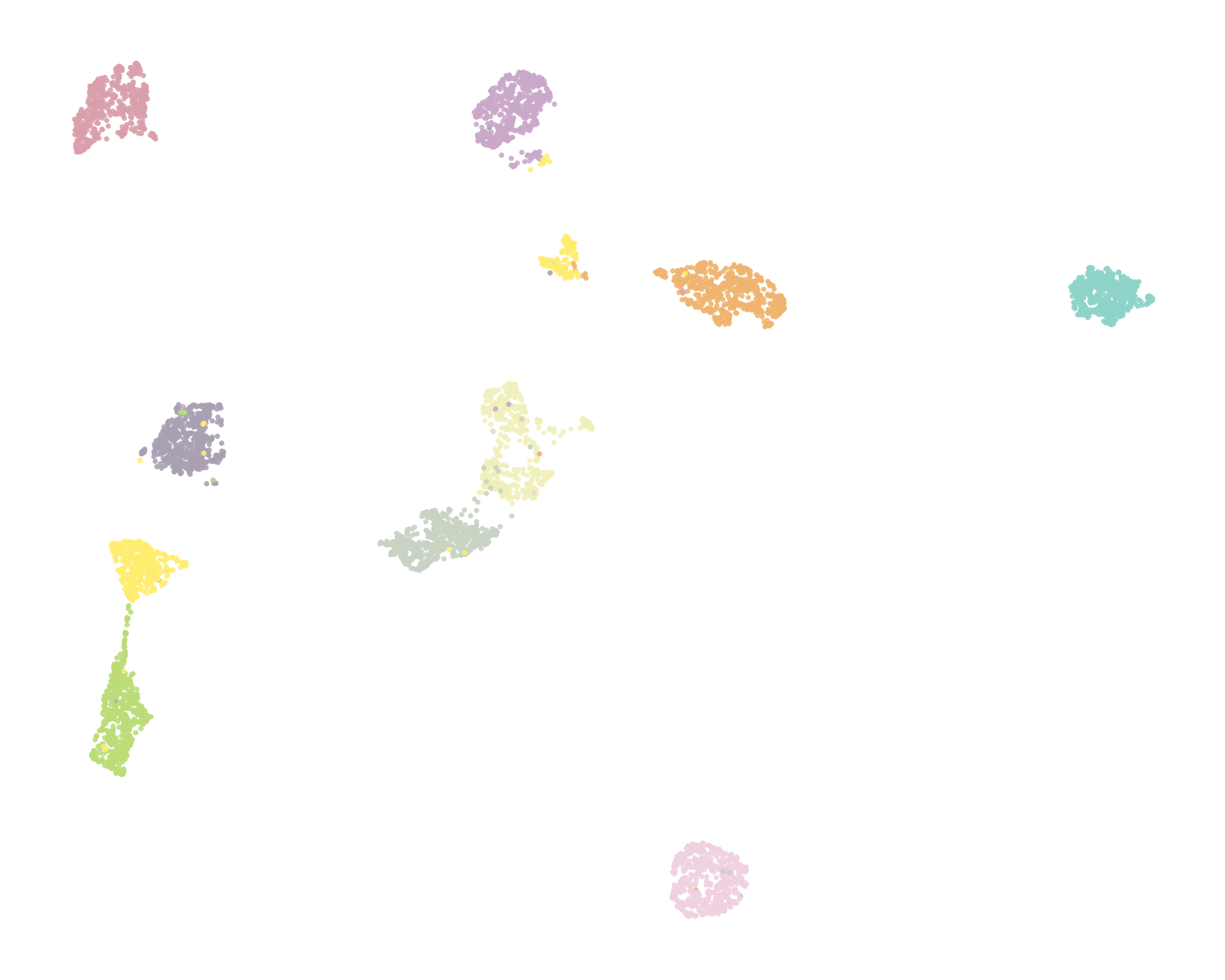}
      \subcaption{Umap, $7.5$s}
    \end{subfigure}
    \begin{subfigure}[b]{0.23\textwidth}
     \includegraphics[width=\textwidth]{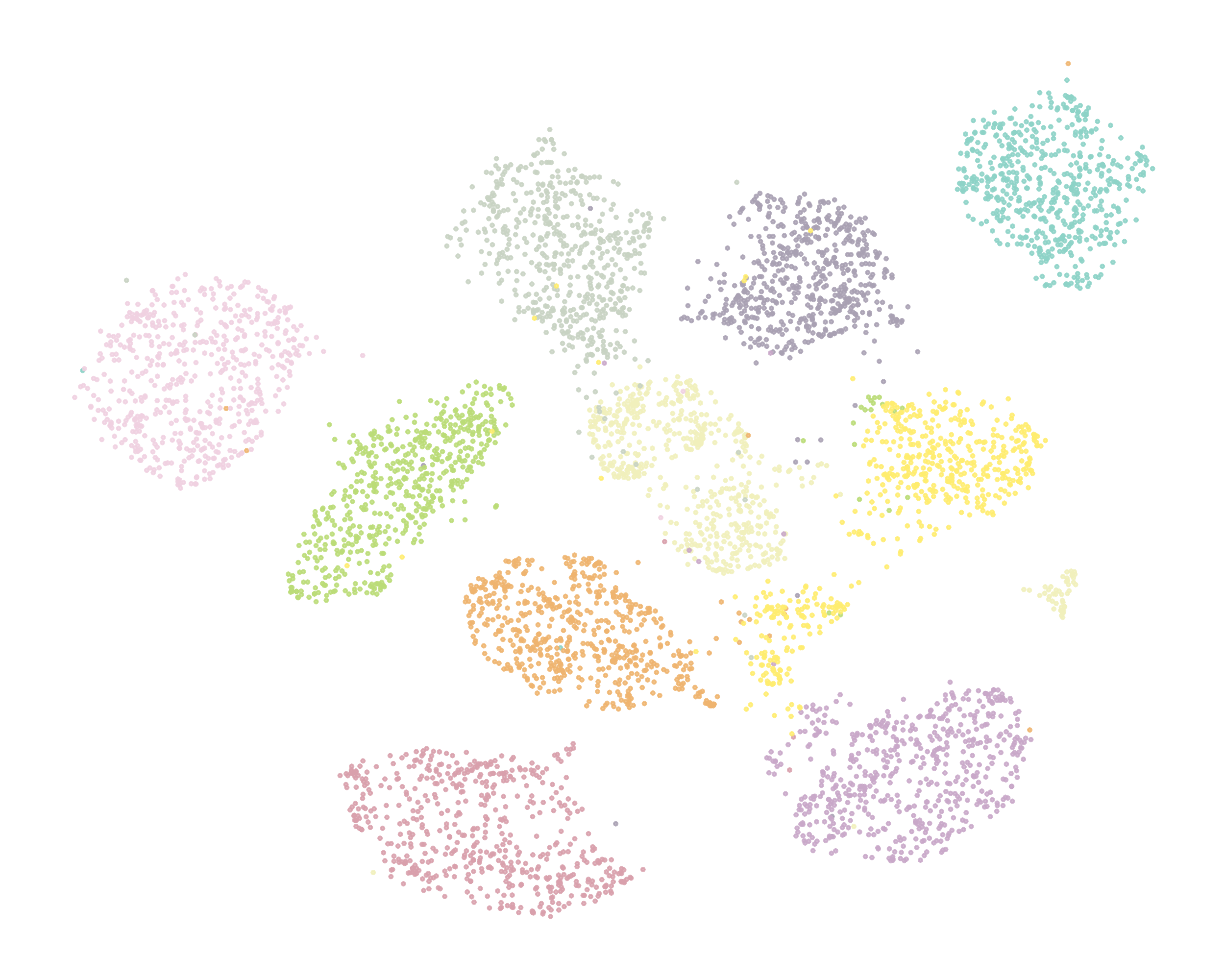}
     \subcaption{NCVis, $0.9$s}
    \end{subfigure}
    \caption{The comparison of visualization methods on handwritten digit data set. NCVis behaves similarly to t-SNE and does not tend to collapse clusters, though reduces the time significantly. Each colored cluster corresponds to a distinct digit.}
    \label{fig:quality}
  \end{figure}

  Our next step was to evaluate the quality of the method and, what is more important, its predictable behavior on simple data sets. We used the Optical Recognition of Handwritten Digits Data Set from~\cite{Dua:2019} which comprised $5620$ preprocessed handwritten digits and thus has a simple structure that is assumed to be revealed by visualization. Figure~\ref{fig:quality} proves that NCVis shows the behavior consistent with classical methods like t-SNE while producing visualization up to the order of magnitude faster.

\subsection{Parallelization Efficiency}
\begin{figure}[t]
    \includegraphics[width=0.46\textwidth]{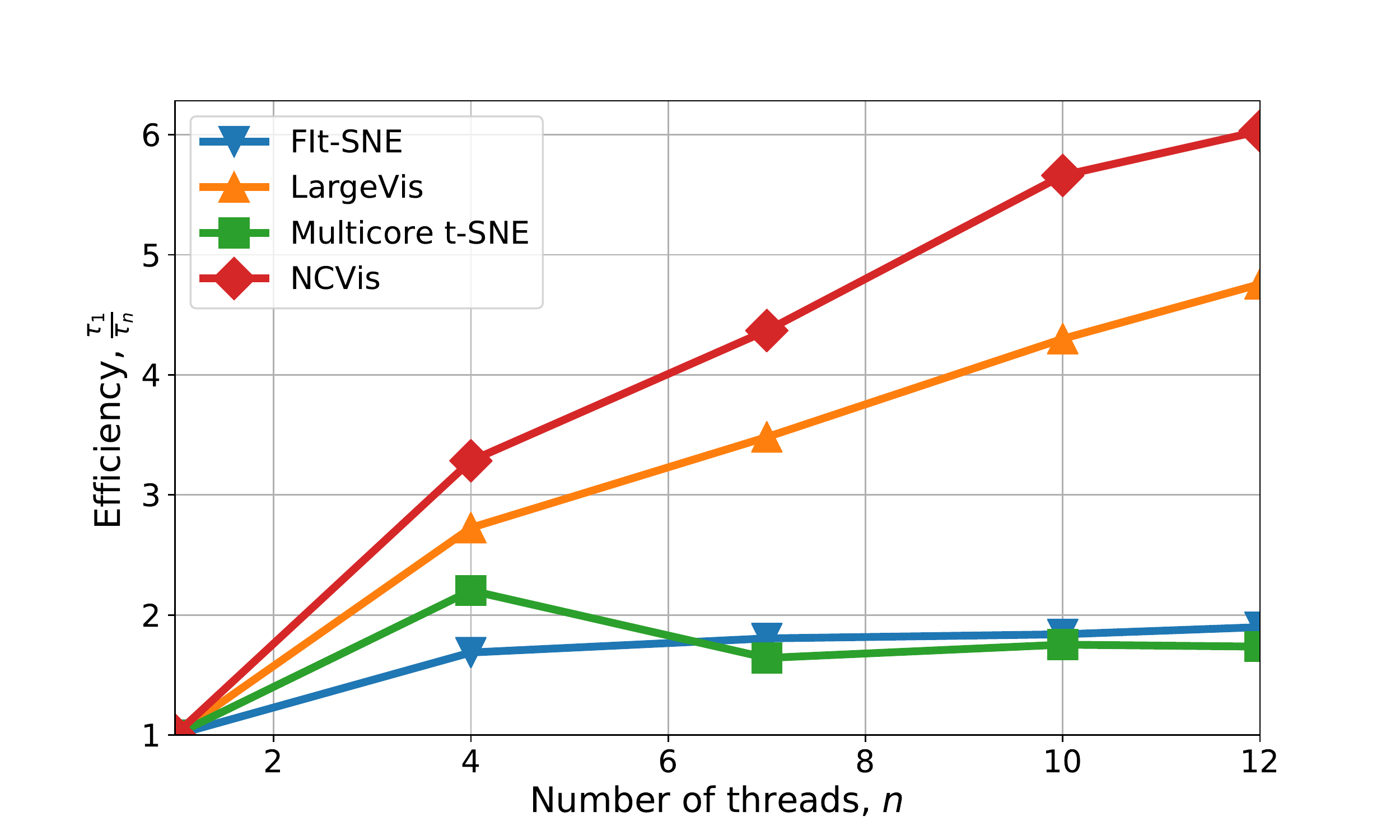}
    \caption{The comparison of visualization approaches by efficiency. Ideally, the efficiency should be equal to the number of threads. NCVis does not achieve this limit but significantly outperforms other methods.}
  \label{fig:efficiency}
  \end{figure}

  Another important criterion to measure the method's scalability is an efficient use of the resources. We can compare the methods, which support parallelization, by the efficiency of parallelization. We use a traditional notation: let $\tau_1, \tau_2, \dots, \tau_n$ be the times which correspond to execution with $1, 2, \dots, n$ processes. One can define efficiency as $S_n = \frac{\tau_1}{\tau_n}$: it equals $1$ for $n=1$ and for more processes we expect $S_n$ to be close to $n$. Even though in practice the achievement of such an efficiency is challenging, we can still use it as a reasonable comparison criterion. The results of the experiment are presented in Figure~\ref{fig:efficiency}. We used $10000$ samples from a news data set. It should be pointed out, however, that slow data loading procedures impact FIt-SNE and LargeVis performance. 

\section{Large Scale Application}
  There are numerous artificial tasks which exhibit a clear cluster structure. On the contrary, pure clusters rarely occur in real-life applications. Nevertheless, there is still inner structure which can be revealed by visualization. To show the full potential of NCVis we will visualize a large India News headlines data set of more than $2$ million samples~\cite{NewsIndia}. The challenge is to study the incredibly vast set of topics in a reasonable amount of time.

\subsection{Data Preprocessing}
  The samples in the data set~$\mathcal{D} = \{s_i\}_{i=1}^N$ are India News headlines. Our task is to build vector representation for each of them. In order to do so, we use $1$ million English word vectors~\cite{mikolov2018advances} trained on Wikipedia 2017, UMBC webbase corpus and statmt.org news data set. These data allow us to perform mapping $\mathcal{R}\colon \mathcal{W} \to \mathbb{R}^M$ for words $w$ in dictionary $\mathcal{W} = \{w_l\}_{l=1}^L$. Each headline~$s_i$ is split into the words~$\{v_k\}_{k=1}^{K_i}$. For each word we then find the representation according to the rule 
  \begin{EQA}[c]
    v_k \to \vu_k =
    \begin{cases}
      \mathcal{R}(v_k), & \text{ if } v_k \in \mathcal{W}, \\
      \bm{0}, & \text{ if } v_k \not \in \mathcal{W}.
    \end{cases}
  \end{EQA}
  Finally, each sentence is represented by the mean of its words:
  \begin{EQA}[c]
    s_i \to \widehat{\vz}_i = \frac{1}{K_i} \sum_{k=1}^{K_i} \vu_k.
  \end{EQA}

\subsection{Data Visualization}
\begin{figure}[t]
    \includegraphics[width=0.46\textwidth]{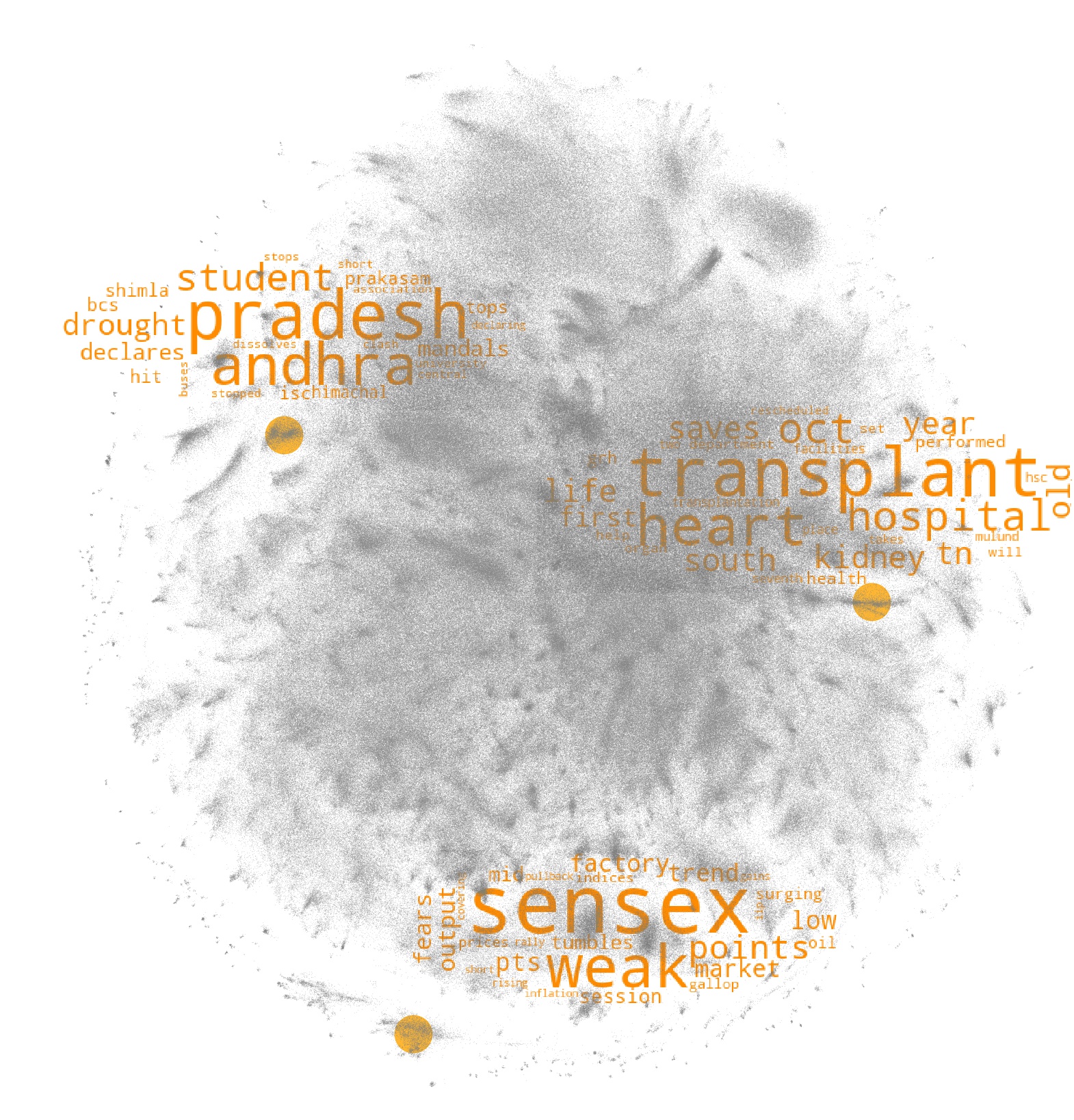}
    \caption{NCVis visualization of India News headlines embeddings. The focus is on several dense and isolated areas, as we believe to find the most interesting information there.}
    \label{fig:isolated}
  \end{figure}

  Afterwards, NCVis is used with cosine similarity as distance function to construct the visualization: 
  \begin{EQA}[c]
    \{\widehat{\vz}_i \in \mathbb{R}^{M}\}_{i=1}^N \to \{\vz_i \in \mathbb{R}^{2}\}_{i=1}^N.
  \end{EQA}
  Cosine similarity was selected as representations for similar words have greater dot product than for dissimilar ones. Thus we assume that the same also holds for sentence representations. The result can be found in Figure~\ref{fig:isolated}.

\subsection{Visualization Analysis}
  There are more than $1$ thousand categories provided in the data set, which makes a standard cluster analysis complicated. However, it can be much more useful to introduce a less excessive list of topics based on the data set structure. One can focus on specific groups of points to develop a better understanding of the presented headlines. We propose to visualize vector representations of news headlines and then to focus on areas of interest: isolated regions and regions with a higher density, see several examples on Figure~\ref{fig:isolated}.

  Within a short time, we have been able to find relevant news topics and study them more carefully by sampling the headlines from one of the selected areas (for example, the one connected with medicine). Several headlines sampled from this ``medicine'' region are presented in Table~\ref{tab:medicine}. We clearly see strong semantic links between the selected headlines which shows that the obtained visualization is meaningful.

  \begin{table}
    \caption{Examples of news headlines selected from the area connected with medicine.}
    \label{tab:medicine}
    \begin{center}
      \begin{tabular}{c}
        \hline
        kidney transplant saves life of year old\\
        \hline
        south tn s first ever heart transplant performed in grh\\
        \hline
        seventh heart transplant takes place in mulund hospital\\
        \hline
        two hsc papers of oct rescheduled to oct\\
        \hline
        two get new shot at life after kidney transplants\\
        \hline
      \end{tabular}
    \end{center}
  \end{table}

\section{Related work}
  The problem of data visualization is well studied and can be solved in several ways.

  The Scikit-learn implementation of t-SNE~\cite{scikit-learn} is probably the most well-known and most widely used visualization method. Among the advantages of the method making it so popular are a convenient, user-friendly, interface and its ability to provide state-of-the-art visualization quality for small and medium-sized data sets. Still, it fails to be efficient enough for large-scale applications because of the normalization problem discussed in Section~\ref{sec:prob}. 

  The issue of scalability can be addressed directly by using the parallel version of the algorithm. Despite the fact that the author of Multicore t-SNE~\cite{Ulyanov2016} adopted this approach and provided the corresponding implementation, the method remains considerably slow and has a low parallelization efficiency, since the normalization term is computed explicitly.

  FIt-SNE~\cite{linderman2019fast} is much more scalable and uses an entirely different idea to compute interactions between the points. It originates from a fast simulation of $N$-body problem, where Fast Fourier Transform can be used to gain a significant speedup. However, only computational part of the algorithm has been changed, and it still behaves similarly to t-SNE in terms of processing large samples.

  All the techniques described above leave the theoretical basis unchanged and try to solve the performance difficulties of the original t-SNE algorithm. Another family of methods takes an entirely different path by introducing its own objective functions. As far as we know, LargeVis~\cite{tang2016visualizing} is the first method with a focus on graph visualization that uses the idea of negative sampling. This vital step removes the necessity of explicit calculations of the normalization term.

  Umap~\cite{mcinnes2018umap} approach is based on the same idea as LargeVis, but it pays much more attention to heuristics, providing a noticeable performance gain. It is also the first method that combines a convenient user interface and efficiency. However, due to the absence of a parallelism support, it faces significant scaling difficulties.

\section{Conclusion}
  We developed NCVis~-- a highly-scalable method for visualization via dimensionality reduction based on the sound theory of Noise Contrastive Estimation. The main idea behind this approach is to focus not only on likelihood maximization between the real data and the model but to keep the model as distinct from the noise as possible. It allows to remove excess information about the object and, at the same time, highlight relevant details. We derive our method from the theory of Noise Contrastive Estimation and outline some of its essential properties. We emphasize that when the amount of noise is large enough, NCVis produces the results, which are close to those obtained with the well-known t-SNE method.

  The experimental validation of NCVis was performed on the conventional multi-core hardware against the state-of-the-art methods for visualization via dimensionality reduction such as Multicore t-SNE~\cite{Ulyanov2016}, Umap~\cite{mcinnes2018umap}, t-SNE implemented in Scikit-learn~\cite{scikitlearn}, FIt-SNE~\cite{linderman2019fast} and LargeVis~\cite{tang2016visualizing}. We show that the proposed method significantly outperforms the competitors in terms of speed: it needs only $6$ minutes to provide well-interpretable visualizations for more than $1$ million of high-dimensional data points. Moreover, we show that the parallelization efficiency of NCVis is much higher than the one of other methods. This fact creates the possibility for visualization of the data of an arbitrary scale given sufficiently powerful hardware. Finally, on the classical data set of handwritten digits~\cite{Dua:2019}, NCVis provides high-quality visualization which is consistent with the famous t-SNE results while computing it 30 times faster.

  The whole pipeline can be found online\footnote{https://github.com/stat-ml/ncvis-examples}. Taking into account that NCVis is easy to install and cross-platform, our method can be considered as a perfect candidate for real-life applications.

\bibliographystyle{ACM-Reference-Format}
\bibliography{www2020}


\end{document}